\documentclass{article}

\usepackage{microtype}
\usepackage{graphicx}
\usepackage{booktabs} % for professional tables
\usepackage{amssymb}
\usepackage{amsmath}
\usepackage{mathtools}
\usepackage{subcaption}
\usepackage{silence}
\usepackage{hyperref}

\DeclareMathOperator*{\argmin}{arg\,min} 

\usepackage[accepted]{icml2018_ida}

\icmltitlerunning{Deep PET Reconstruction}

\newcommand{\E}[1]{E$^{\text{#1}}$}

\begin{document}

\twocolumn[
\icmltitle{DeepPET: A deep encoder--decoder network for directly solving the PET reconstruction inverse problem}
%



\icmlsetsymbol{equal}{*}

\begin{icmlauthorlist}
\icmlauthor{Ida H{\"a}ggstr{\"o}m}{mskcc}
\icmlauthor{C.~Ross Schmidtlein}{mskcc}
\icmlauthor{Gabriele Campanella}{mskcc,cornell}
\icmlauthor{Thomas~J. Fuchs}{mskcc,cornell,mskcc2}
\end{icmlauthorlist}

\icmlaffiliation{mskcc}{Department of Medical Physics, Memorial Sloan Kettering Cancer Center, New York, NY}
\icmlaffiliation{cornell}{Department of Physiology and Biophysics, Weill Cornell Medicine, New York, NY}
\icmlaffiliation{mskcc2}{Department of Pathology, Memorial Sloan Kettering Cancer Center, New York, NY}

\icmlcorrespondingauthor{Ida H{\"a}ggstr{\"o}m}{haeggsti AT mskcc DOT org}

\icmlkeywords{Machine Learning, ICML}

\vskip 0.3in
]



\printAffiliationsAndNotice{}  % leave blank if no need to mention equal contribution
\begin{abstract}
%stage
Positron emission tomography (PET) is a cornerstone of modern radiology. The ability to detect cancer and metastases in whole body scans fundamentally changed cancer diagnosis and treatment.

%problem
One of the main bottlenecks in the clinical application is the time it takes to reconstruct the anatomical image from the deluge of data in PET imaging. State-of-the art methods based on expectation maximization can take hours for a single patient and depend on manual fine-tuning. This results not only in financial burden for hospitals but more importantly leads to less efficient patient handling, evaluation, and ultimately diagnosis and treatment for patients.

%solution
To overcome this problem we present a novel PET image reconstruction technique based on a deep convolutional encoder--decoder network, that takes PET sinogram data as input and directly outputs full PET images. Using realistic simulated data, we demonstrate that our network is able to reconstruct images $>$100 times faster, and with comparable image quality (in terms of root mean squared error) relative to conventional iterative reconstruction techniques. 

% {\color{blue}This document provides a basic paper template and submission guidelines.
% Abstracts must be a single paragraph, ideally between 4--6 sentences long.
% Gross violations will trigger corrections at the camera-ready phase.}
\end{abstract}
%%%%%%%%%%%%%%%%%%%%%%%%%%%%%%%%%%%%%%%%%%%%%%%%%%%%%
\section{Introduction}
\label{sec:introduction}
Positron emission tomography (PET) is widely used for numerous clinical, research, and industrial applications, due to its ability to image functional and biological processes \emph{in vivo}. Of all these applications, PET's integral role in cancer diagnosis and treatment is perhaps its most important. PET is able to detect radiotracer concentrations as low as picomolar. This extreme sensitivity enables earlier and more precise diagnosis and staging. Early diagnosis of cancer is strongly correlated with outcome, as is early intervention in treatment management, where ineffective treatments can be stopped and more effective ones substituted in their place. All benefits of PET in cancer care strongly rely on PET image quality however, making it a necessity with reliable, quantitative, high quality images.   

In PET, the subject is injected with a tracer labeled with a positron emitting radionuclide. The physical decay of the nuclide will lead to the emission of two nearly back-to-back photons emanating from a positron/electron annihilation event. The raw PET tomographic data is typically stored in projection data histograms called \emph{sinograms} (example in Figure \ref{fig:autoencoder}), where each element in the sinogram represents the number of photon coincidence detections in a particular detector pair. As a result, these detector pairs can be thought of as representing projections of the underlying activity distribution onto planes perpendicular to the lines connecting the detector pairs.

Sinogram data cannot be directly interpreted by an observer, but can be shown via the projection-slice theorem that they map to images. However, random process noise in the data makes this relationship ill-posed, and the reconstruction of the tracer distributions function can be solved as an inverse problem.
This is known as image reconstruction in the tomography community and is used in this sense in this paper. The inverse problems for PET image reconstruction represent explicit models of the mean values of the physical system, where a particular model's inverse represents a solution to the image reconstruction problem. 
% Unfortunately, the data that these models use are from random processes, which lead to data/model inconsistency and an ill-posed inverse problem.
% Because of this a number of simplifying assumptions are made to make the problem tractable. These include the use of a discrete linear system to model the data to image mapping, simplified statistical models to account for the degree of data/model inconsistency, and regularization to control the data overfitting problem.

Various PET image reconstruction techniques exist, the most common ones being analytical filtered back-projection (FBP), and maximum likelihood (ML) methods such as maximum-likelihood expectation maximization (MLEM) or its incremental update version, ordered subset expectation maximization (OSEM). Typically however, images resulting from these standard methods suffer from data/model mismatches, data inconsistency, and data overfitting, which can manifest as artifacts such as streaks and noise in the reconstructed images. As a result, the reconstructed images are generally filtered at the end of the reconstruction process. In the case of ML methods, regularization can be used to overcome the ill-posedness and reduce the fitting noise in the reconstructed images. However, while regularization for PET image reconstruction has been around for a while, regularization requires a large number of iterations to make its benefits apparent, and as a result has only recently been used clinically. 
% Nonetheless these methods have been very successful in providing physicians with crucial information concerning patient state, or researchers with better insight into the molecular properties of biological systems.

While these image reconstruction methods have been successful there are still some aspects of the process that can be improved. 
% Both the system and statistical models contain a number of simplifications that contribute to the system model mismatches. These include the simplified photon/electron (positron) transport models, simplified geometry, and the neglect of other noise sources, such as dead time. 
Regularization in particular is an open problem in PET image reconstruction. Many regularization models have been proposed \cite{Sidky2012,Teng2016,Schmidtlein2017} but there is no clear consensus on how to choose between them. The use of a deep network carries some advantages. It at once can be trained to learn the solution to the inverse problem in the presence of noise. That is, it learns the inverse of the physical model, the appropriate statistical model, and the regularization that best fits the character of the data.  

Another advantage of a using a deep network for reconstruction is its potential for computational efficiency. The PET inverse problem is generally cast as a minimization problem and solved via a gradient algorithm scheme (or something similar). At each step of the descent at least two multiplications by the PET system matrix is required. The PET system matrix is very large and these steps are expensive to compute, especially when performing multiple iterations. Viewing the deep network as a regularized inverse to the PET system model one can envision it as performing a single one of these steps, where the step now represents the inverse operator. Hence, we propose an encoder--decoder network that uses supervised learning to solve the PET reconstruction inverse problem directly.

%%%%%%%%%%%%%%%%%%%%%%%%%%%%%%%%%%%%%%%%%%%%%%%%%%%%%
\section{Related work}
\label{sec:relatedwork}
In recent years, deep learning has shown great potential in many medical image restoration, segmentation, and analysis applications. In particular, encoder--decoder architectures have been readily used for these purposes. Convolutional encoder--decoder (CED) models are capable of stepwise compressing the input image data into a latent space representation, and then stepwise rebuild that representation into a full dataset. CEDs have been used to restore low dose computed tomography (CT) images to full dose images \cite{Chen2017}, and to generate synthetic CT images from MR images used for attenuation correction in PET/MR \cite{Liu2017}. \cite{Yang2017} proposed an artificial neural network to determine optimal fusing of three different maximum \emph{a posteriori} (MAP) PET reconstructions of different regularizing weights, and \cite{Cui2017} proposed a stacked sparse autoencoder to improve the quality of dynamic PET MLEM reconstructions. \cite{Jin2017} investigated the use of deep convolutional neural networks (CNNs) to solve inverse problems in imaging, and used their network to regress sparse view CT FBP images to full view images. 

All the works above are post processing methods using reconstructed images as network input to restore image quality and reduce image defects and noise. It is less explored how to use deep learning methods within the PET image reconstruction process itself, i.e. as part of generating PET images directly from PET sinogram data. 
\cite{Chen2017a} used a 3 layer CNN to optimize the regularizing term in iterative CT reconstruction. Although the CNN is part of the reconstruction loop, their method still resembles a conventional regularized iterative least squares approach where the system matrix has to be known. Also using the concept of iterative CNN reconstruction, \cite{Gong2017} proposed a residual convolutional autoencoder within an ML framework to denoise PET images. In a recent approach similar to the work presented here, \cite{Zhu2018} studied image reconstruction for various medical imaging modalities by deep learning the transform between sensor and image domain. Although their work focuses on MRI data, PET reconstruction is mentioned using a single example.

The aim of this work is to design a deep CED architecture that directly and quickly reconstructs PET sinogram data into high quality images. To our best knowledge, this will be the first systematic work of its kind in the new field of direct deep learning-based tomographic image reconstruction.

%%%%%%%%%%%%%%%%%%%%%%%%%%%%%%%%%%%%%%%%%%%%%%%%%%%%%
\section{Theory}
\label{sec:theory}
%%%%%%%%%%%%%%%%%%%%%%%%%%%
\subsection{Emission computed tomography image reconstruction}
\label{sec:convrecon}
Emission computed tomography (ECT) image reconstruction is based on a Poisson noise model given by
%------------------------------
\begin{equation}
g = \text{Poisson}\{Af+\gamma\},
\label{eqn:PET}
\end{equation}
%------------------------------
where $g$ is the measured sinogram data, $A$ is the linear projection operator, $f$ is the unknown activity distribution to be estimated, and $\gamma$ is the additive counts from random and scatter events.
% where $g\in\mathbb{N}^{N}$ is the measured projection data, $A\in\mathbb{R}^{N \times M}$ is the linear projection operator, $f\in\mathbb{R}_{\text{+}}^{M}$ is the unknown activity distribution to be estimated, and $\gamma \in\mathbb{R}_{\text{+}}^{N}$ is the additive counts from random and scatter events. 
%%%%%%%%%%%
% This model can be solved by minimizing the residual of KL-divergence of the data model, which results in the minimization problem given by
% %------------------------------
% \begin{equation}
% f = \argmin_{f}\left\{ \langle A f,1 \rangle - \langle \log{ A f +\gamma} ,g \rangle\right\},
% \label{eqn:PLmin}
% \end{equation}
% %------------------------------
% where $ \langle \cdot, and \cdot \rangle$ is the inner product. In this form the physical aspects of the model (e.g., geometry and statistical distribution) are accounted for by the KL-divergence term.
This model can be solved by minimizing the residual of KL-divergence of the data model and a regularization term, which results in the minimization problem given by
%------------------------------
\begin{equation}
f = \argmin_{f}\left\{ \langle A f,1 \rangle - \langle \log{ A f +\gamma} ,g \rangle + \lambda R(f)\right\},
\label{eqn:PLmin}
\end{equation}
%------------------------------
where $ \langle \cdot, \cdot \rangle$ is the inner product, $R(f)$ is a regularization function, and $\lambda$ the regularization weight. In this form the physical aspects of the model (e.g., geometry and statistical distribution) are accounted for by the KL-divergence term.  
However, the model in Equation~(\ref{eqn:PLmin}) contains a number of assumptions. These include the approximation of the geometric relationship between the image and detectors, the associated point spread function, the estimate of the additive count distribution, the validity of the Poisson statistical model as applied to the actual data, and the functional form of the regularizer. 
In particular, the optimal regularization function is not generally known and is an active area of research \cite{Sidky2012,Teng2016,Schmidtlein2017}. 
% As a result, while the mathematical description in Equation~(\ref{eqn:PLmin}) provides a well defined functional form, it relies on the assumption that its constituent components are good approximations of the true physical system.  Poor approximations of these components may reveal themselves as data/model inconsistency, but this provide little insight into how to improve them.

On the other hand, a deep learning network has the potential to learn each of these aspects of the model from the data itself. The network learns the correct geometric, statistical, and regularization models, provided the training examples are realistic. As a result, knowledge of all these properties do not need to be explicitly included in the network model. However, both the geometry and statistical properties may still be implicitly included through the model used to create the training data. In this study, because these networks can learn rather than using defined statistical properties of the data, we use precorrected data, where the $i^{\text{th}}$ precorrected data element is given by 
%------------------------------
\begin{equation}
\hat{g}_i = \frac{g_i-\gamma_i}{\mu_i},
\label{eqn:predata}
\end{equation}
%------------------------------
where $\mu_i$ is the attenuation in the $i^{\text{th}}$ coincidence count. We note that precorrecting the data mixes its statistical properties and, as a result, is generally avoided in the inverse problem formulation due to its explicit use of a particular statistical model, Poisson in the case of PET. %{\color{red}Add one sentence, why this is not a (significant) problem in our setup}

%%%%%%%%%%%%%%%%%%%%%%%%%%%%%%%%%%%%%%%%%%%%%%%%%%%%%
\section{Experimental design}
\label{sec:experiments}
We propose a novel PET image reconstruction approach, based on a deep CED network, as opposed to traditional iterative reconstruction techniques.

%%%%%%%%%%%%%%%%%%%%%%%%%%%
\subsection{Dataset description}
\label{sec:simulations}
The major bottleneck in many deep learning experiments is the limited size of available datasets and lack of labeled data. We circumvented this problem by generating labeled data synthetically. We used the open-source PET simulation software PETSTEP \cite{Berthon2015,Haggstrom2016} to simulate PET scans and generate realistic PET data that have been validated by GATE \cite{Jan2004} Monte Carlo (MC). They include effects of scattered and random coincidences, photon attenuation, counting noise, and image system blurring. We modeled a GE D710/690 PET/CT scanner, with sinogram data of 288 angular$\times$381 radial bins. 
This approach avoids too simplistic training data, for example by only adding Poisson noise to sinogram data, and thus enables the transferability of the trained network to clinical data. 

We used the OSEM reconstructed 2D image slices of 128$\times$128 pixels (700~mm field of view) from real PET whole-body patient scans as phantom input to PETSTEP, along with the corresponding size-matched CT images used for attenuation information. A total of 79 patients with on average 256 slices (whole-body scans) each were used.
% where about 12 random activity levels of each image slice was simulated, resulting in 111,935 projection datasets (total, trues, scatters, randoms, and attenuation factors $\mu$). The noisy total projection data had on average 1.4$\cdot10^{6}$ total counts (range 3.7$\cdot10^{3}$ to 3.4$\cdot10^{7}$).
% The original 111,935 images were used as ground truth, and the simulated total projection data was used as network input after being precorrected for randoms, scatters and attenuation according to Equation~(\ref{eqn:predata}). The precorrected projection data was also cropped prior to input. The image field of view is circular with the same diameter as the image matrix size (here 128), leaving corners empty. This in turn leaves the first and last 56 radial projection data bins empty, which were thus cropped to reduce the number of network elements.making a total of 200$\cdot$280=56,000 unique 2D activity images (with attenuation $\mu$-maps).
PET acquisitions of these phantom images were simulated using PETSTEP, where the random and scatter fraction was randomly drawn from a normal distribution around realistic values for the given activity level and object size.
Nine random activity (noise) levels, from low to high, of each image slice was simulated, resulting in sinogram datasets containing total, trues, scatters, randoms, and attenuation factors. For 3 of the 9 realizations, the phantom image was randomly translated ($\pm25$ pixels in $x$ and $y$-dir) and rotated ($\pm$10$^\circ$), and one of those 3 realizations was also left-right flipped, prior to simulation. Sets with a noisy total count of $<$1\E{5} or $>$5\E{7} were discarded. 176,585 sinogram datasets were kept, and the noisy total sinograms had on average 5.4\E{6} total counts.
The original 176,585 phantom images were used as ground truth, and the simulated total sinogram data was used as network input after being precorrected for randoms, scatters and attenuation according to Equation~(\ref{eqn:predata}). The precorrected sinograms were also cropped prior to input. The image field of view is circular with the same diameter as the image matrix size (here 128), leaving corners empty. This in turn leaves the first and last 56 radial sinogram bins empty, which were thus cropped to reduce the number of network elements.

%%%%%%%%%%%%%%%%%%%%%%%%%%%
\subsection{Convolutional encoder-decoder design}
\label{sec:encoderdecoder}
%------------------------------
\begin{figure*}[ht!]
\vskip 0.2in
\begin{center}
\centerline{\includegraphics[width=\textwidth]{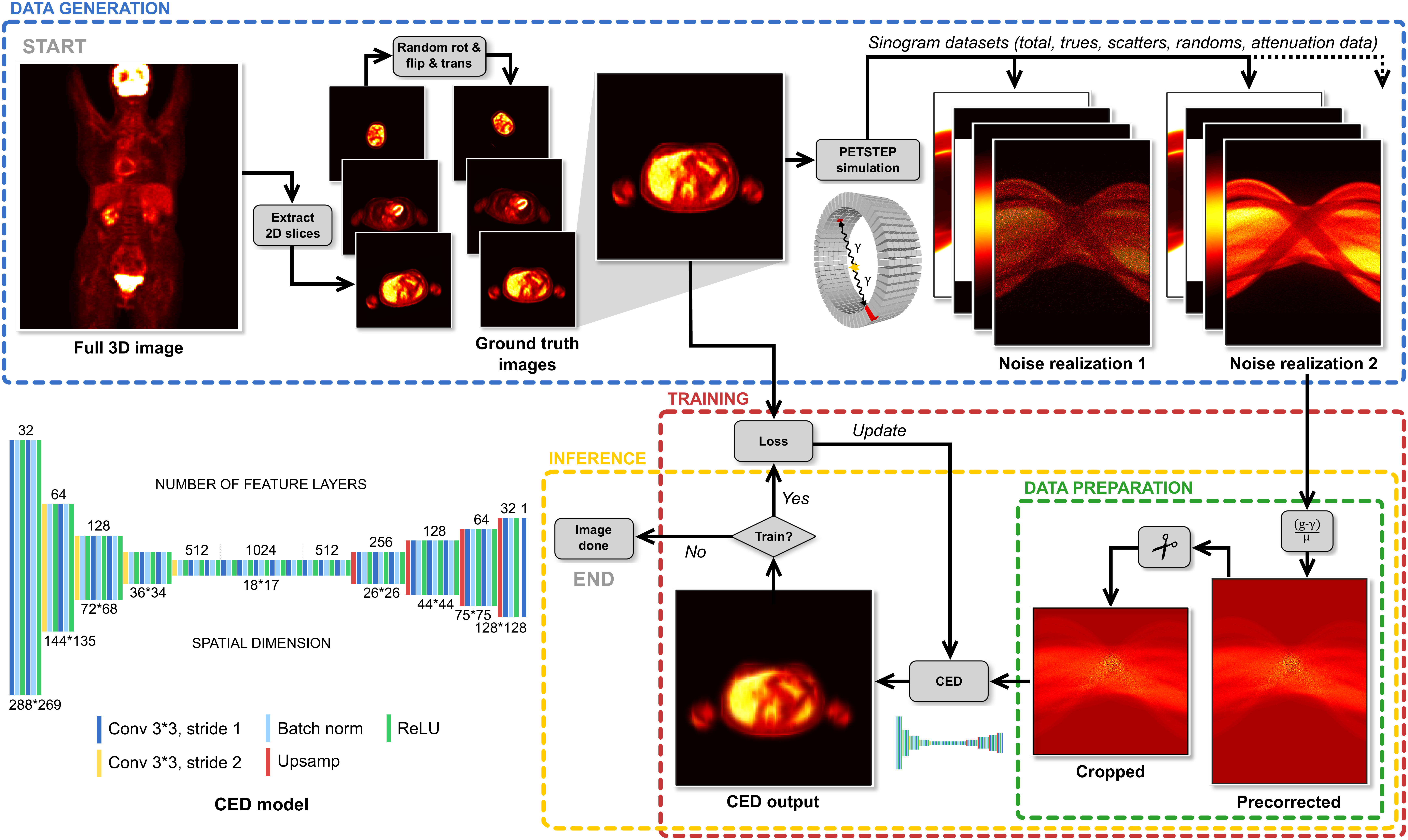}}
\caption{Schematic illustration of the reconstruction pipeline and the convolutional encoder--decoder (CED) architecture. The data generation process is depicted on the top, including simulation with PETSTEP, resulting in multiple sinogram datasets at different noise levels. Experimental setup of the deep learning training and inference procedure is shown on the bottom right, and the CED design on the bottom left.}
\label{fig:autoencoder}
\end{center}
\vskip -0.2in
\end{figure*}
%------------------------------
The encoder of the final model loosely mimics the VGG16 network architecture \cite{Simonyan2014} with modifications, and is depicted in Figure~\ref{fig:autoencoder}. 
The sinogram input data is of size 288$\times$269$\times$1, and the output in image space is of size 128$\times$128$\times$1.
The encoder contracts the input data in a manner typical to CNNs. It consists of sequential blocks of 3$\times$3 convolutions with stride 2 and a factor 2 increase in the number of output feature layers, followed by batch normalization (BN) and activation by a rectified linear unit (ReLU).
The encoder output consists of 1024 feature maps of size 18$\times$17. Each feature is a non-linear function of an extensive portion of the input sinogram. This is of special interest in this PET scenario since single points in the reconstructed image domain are represented by sinusoidal traces in the input domain. In other words, a large spread out portion of the input image data is needed to infer each reconstructed image pixel.

The decoder upsamples the contracted feature representation from the encoder into PET images. Each step in the decoder path consists of an upsampling layer, increasing the image size by a factor 1.7, a 3$\times$3 convolution that halves the number of feature layers, a BN layer, followed by a ReLU. 

We implemented a number of different CED designs, with different number of convolutional layers and feature layer depths, varying spatial sizes, as well as the Adam stochastic gradient optimization method \cite{Kingma2014}, and evaluated which one resulted in highest quality of reconstructed images. The most relevant models investigated are denoted M1 through M7 as well as the ultimately chosen model named DeepPET, and are seen in Table~\ref{tab:models}.
% \begin{itemize}
% \item M1: 20 conv layers, 256 feature layers, no BN, Adam
% \item M2: 20 conv layers, 256 feature layers, Adam
% \item M3: 23 conv layers, 512 feature layers, Adam
% \item M4: 31 conv layers, 1024 feature layers, Adam
% \item M5: 31 conv layers, 1024 feature layers, SGD
% \item M6: 36 conv layers, 2048 feature layers, Adam
% \end{itemize}
\begin{table}[t]
\caption{Parameterization of the seven most relevant encoder--decoder architectures in comparison to DeepPET.}
\label{tab:models}
\vskip 0.15in
\begin{center}
\begin{small}
\begin{sc}
\begin{tabular}{lcccr}
\toprule
Name & Conv & Feature &  Opt \\
 & layers & layers &  \\
\midrule
M1      & 26 & 256 & SGD \\
M2      & 29 & 256 & SGD \\
M3      & 31 & 512 & SGD \\
M4      & 31 & 512 & Adam \\
DeepPET & 31 & 1024 & SGD \\
M5      & 31 & 1024 & Adam \\
M6      & 36 & 512 & SGD \\
M7      & 36 & 2048 & SGD \\
\bottomrule
\end{tabular}
\end{sc}
\end{small}
\end{center}
\vskip -0.1in
\end{table}

%%%%%%%%%%%%%%%%%%%%%%%%%%%
\subsection{Implementation and training procedure}
\label{sec:training}
The PETSTEP simulated dataset was divided on a patient level into three splits for training (116,065 sinogram datasets, 66\%), validation (24,789 sinogram datasets, 14\%) and testing (35,731 sinogram datasets, 20\%). All three sets were kept separate. 

The network was implemented in PyTorch \cite{pytorch}, and trained on NVIDIA GTX 1080Ti GPUs. The mean squared error (MSE) was used as loss function, 
%adjusted for zero valued pixels. 
%Thus, rRMSE$=\frac{1}{n}\sum_{i=1}^{n}{l_i}$ with 
%------------------------------
\begin{equation}
\text{MSE} = 
\frac{1}{n}\sum_{i=1}^{n}\left(x_i-y_i\right)^{2},
\label{eqn:lossmse}
\end{equation}
%------------------------------
% %------------------------------
% \begin{equation}
% l_i = 
% \begin{dcases}
% \left(\frac{x_i-y_i}{y_i}\right)^{2}, & {y_i>0}\\
% {x_i^{2}}, & {y_i=0}
% \end{dcases}
% \label{eqn:loss}
% \end{equation}
% %------------------------------
where $x$ is the model image, $y$ the ground truth, and $n$ the number of image pixels. The model was optimized with stochastic gradient descent (SGD) with momentum on mini-batches and learning rate decay.
The hyperparameters used for training were: learning rate 0.005; batch size 30; BN momentum 0.2; SGD decay: 1/2 every 20 epochs; SGD momentum 0.9; bilinear upsampling. The model was optimized on the training set over 100 epochs, and the MSE was calculated on the validation set every 5\emph{th} epoch. After finishing training, the model with optimal performance on the validation set was used on the test set. 
%%%%%%%%%%%%%%%%%%%%%%%%%%%
\subsection{Conventional image reconstruction}
\label{sec:conventionalrecon}
Apart from the deep reconstruction method proposed here, the PET sinograms were also reconstructed using conventional techniques; FBP and unregularized OSEM with 5 iterations and 16 subsets. Typically for the GE D710/690, a transaxial 6.4~mm FWHM Gaussian postfilter is used together with a 3 point axial smoothing. Here the data consisted of individual 2D slices, hence only transaxial filtering was used.
%%%%%%%%%%%%%%%%%%%%%%%%%%%
\subsection{Image quality evaluation}
For image quality evaluation, the relative root mean squared error (rRMSE) was used,
%------------------------------
\begin{equation}
\text{rRMSE} = 
\sqrt{\text{MSE}}/\bar{y},
\label{eqn:lossrrmse}
\end{equation}
%------------------------------
where $\bar{y}$ is the ground truth average pixel value.  

%%%%%%%%%%%%%%%%%%%%%%%%%%%%%%%%%%%%%%%%%%%%%%%%%%%%%
\section{Results}
\label{sec:results}
\begin{figure}[tb!]
\vskip 0.2in
\begin{center}
\centerline{\includegraphics[width=\columnwidth]{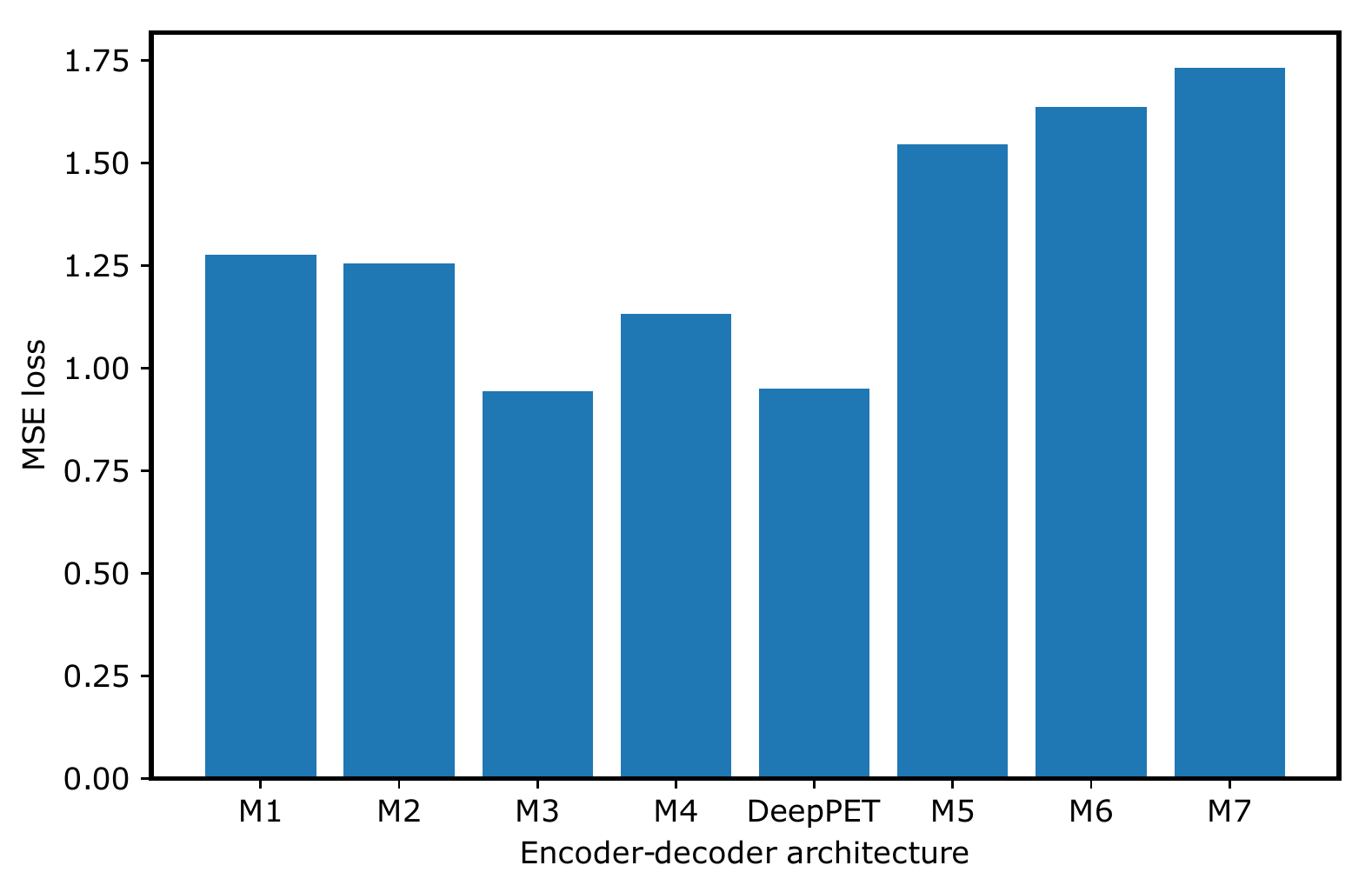}}
\caption{The minimum validation loss up to epoch 60 for the model design M1 through M7, and the chosen model denoted DeepPET. *M7 only had data up to 50 epochs.}
\label{fig:designloss}
\end{center}
\vskip -0.2in
\end{figure}

\begin{figure}[t!]
% \vskip 0.2in
\begin{center}
\centerline{\includegraphics[width=\columnwidth]{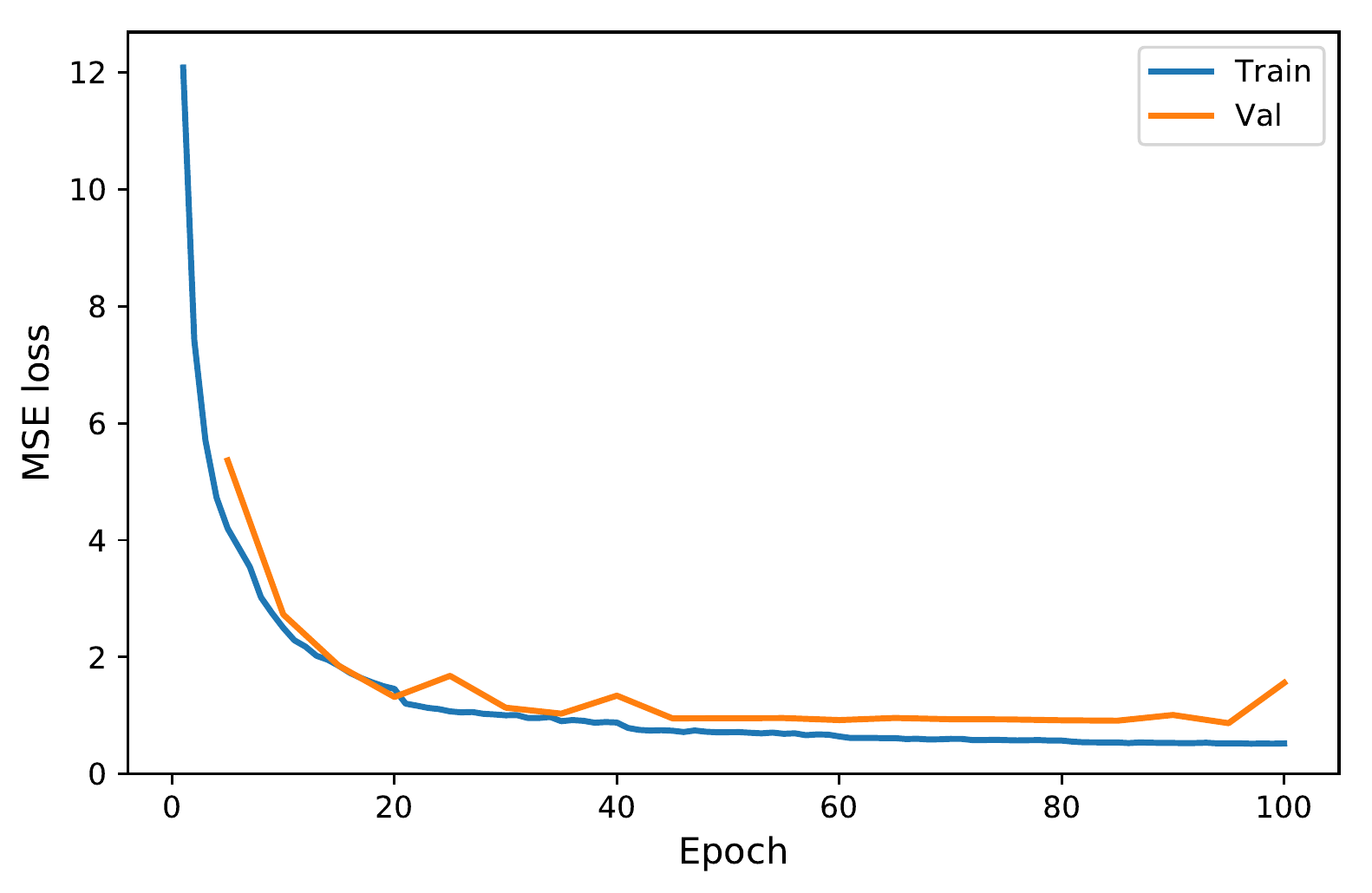}}
\caption{Convergence behavior of average MSE calculated between the ground truth simulation and the reconstructed PET images. Depicted is the training error in blue and validation error in orange for each epoch.}
\label{fig:trainloss}
\end{center}
\vskip -0.2in
\end{figure}

\begin{figure*}[ht]
\vskip 0.2in
\begin{center}
\centerline{\includegraphics[width=1\textwidth]{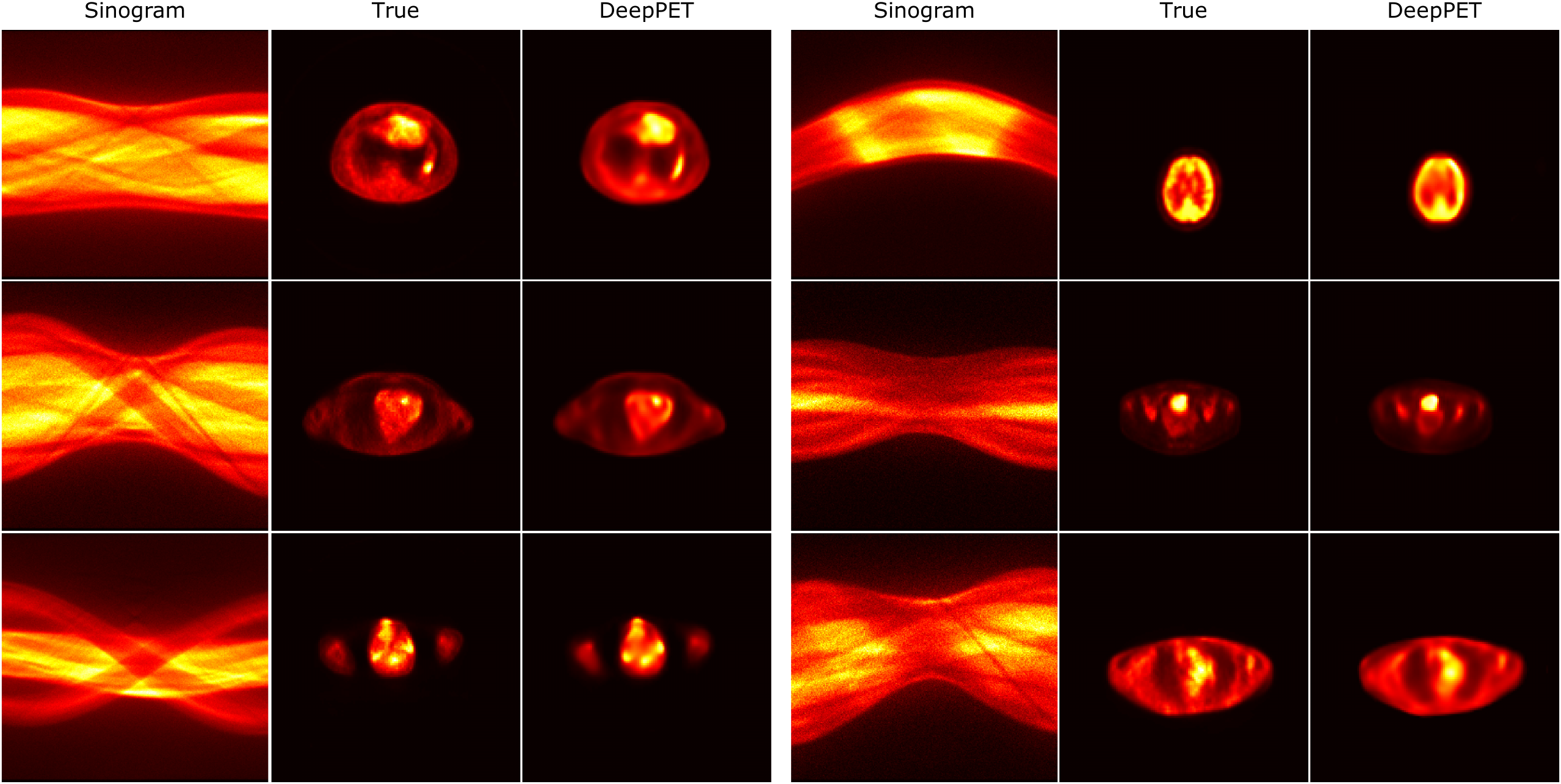}}
\caption{Examples of successful DeepPET reconstructions from the test set. For each of the two grouped columns, left to right: PET sinogram (network input), ground truth, and the DeepPET reconstruction.}
\label{fig:reconsgood}
\end{center}
\vskip -0.2in
\end{figure*}

\begin{figure*}[ht!]
\vskip 0.2in
\begin{center}
\centering
\begin{subfigure}[b]{1\textwidth}
   \centerline{\includegraphics[width=1\linewidth]{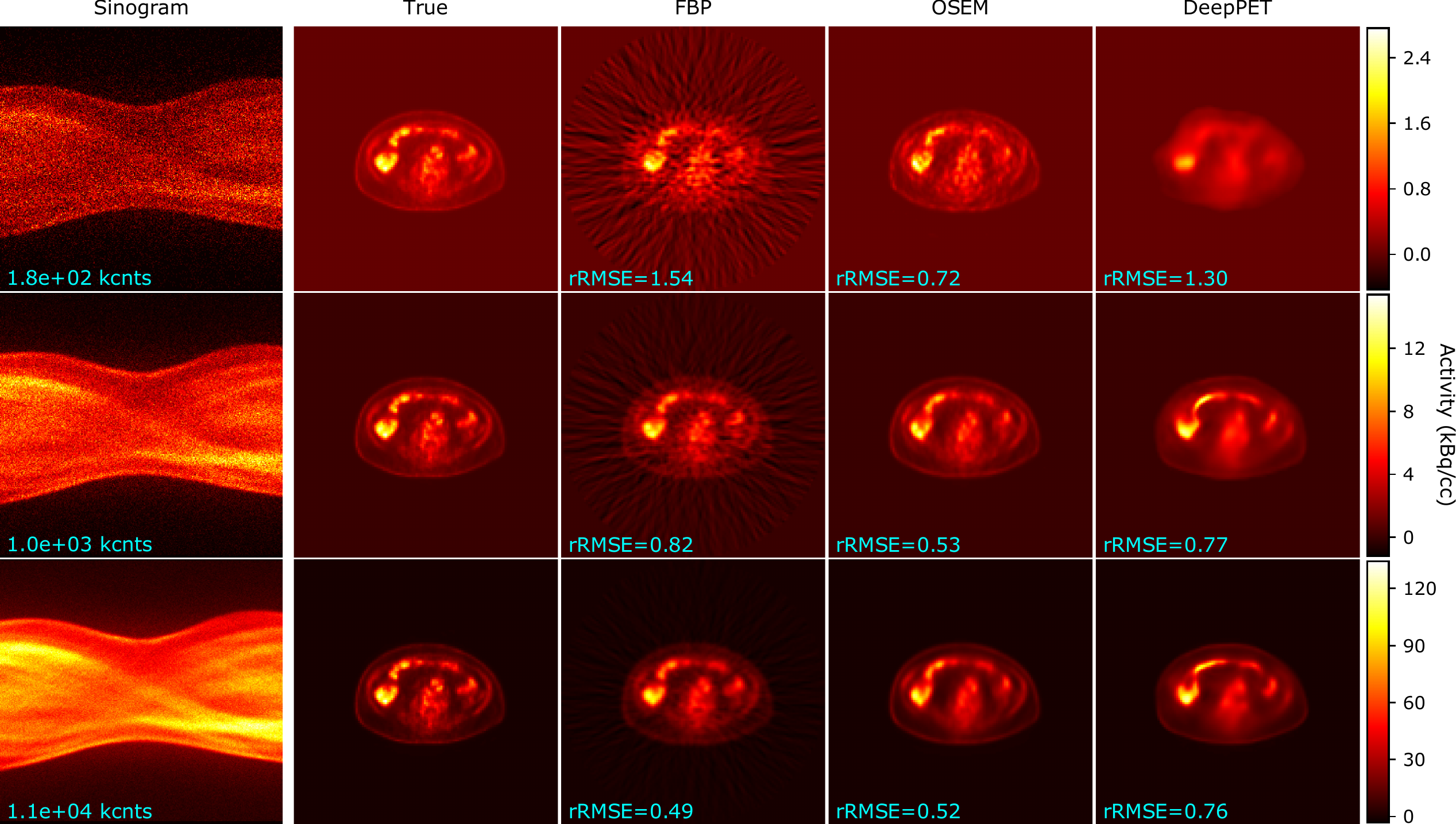}}
   \caption{}
\end{subfigure}
\par\medskip
\begin{subfigure}[b]{1\textwidth}
   \centerline{\includegraphics[width=1\linewidth]{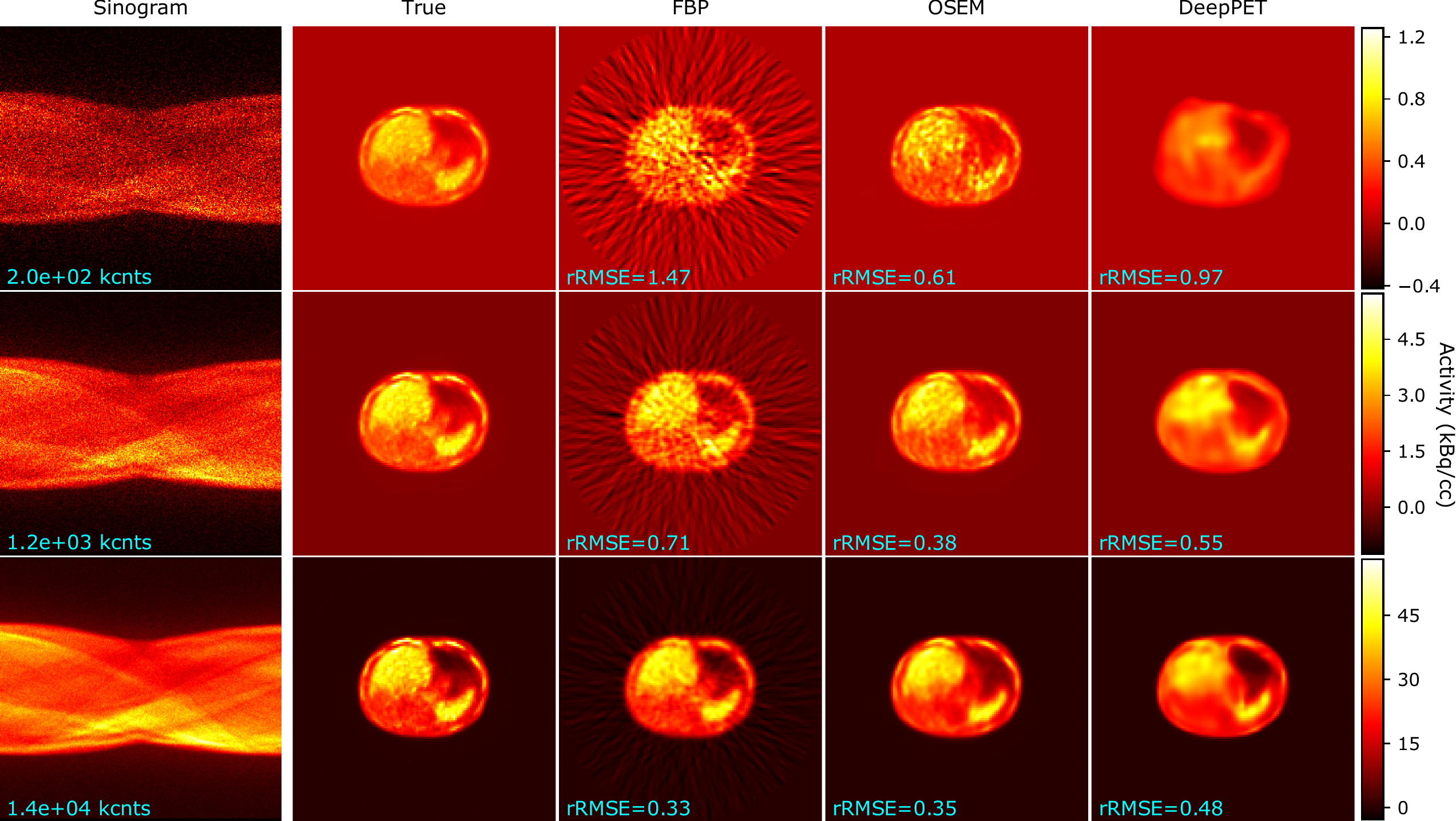}}
   \caption{}
\end{subfigure}
\caption{Examples from the test set at different noise levels. For (a) and (b):  Top to bottom: low count (high noise) to high count (low noise) sinogram data. Left to right: PET sinogram (network input), ground truth, FBP, OSEM, and the DeepPET reconstruction. Sinograms are labeled with their total counts, and images with average rRMSE relative ground truth average.}
\label{fig:recons}
\end{center}
\vskip -0.2in
\end{figure*}

For model designs M1 through M7, as well as DeepPET, the minimum validation set MSE loss up to epoch 60 is seen in Figure~\ref{fig:designloss}.
% , and two reconstructed test set images for the different designs is seen in Figure~\ref{fig:designs}. 
The Adam optimizer was found to yield poorer results compared to SGD. Resulting images lacked detail and were more blurry. Using SGD, the model was able to capture smaller image structures and reduce the loss.
%------------------------------
%------------------------------
% %------------------------------
% \begin{figure*}[ht!]
% \vskip 0.2in
% \begin{center}
% \centerline{\includegraphics[width=\textwidth]{compareModelLoss_imgs.pdf}}
% \caption{Two example images reconstructed with the different model architectures M1 through M6 (trained 60 epochs), where the chosen model is denoted DeepPET. The models yield varying image quality in terms of artifacts and level of captured detail.}
% \label{fig:designs}
% \end{center}
% \vskip -0.2in
% \end{figure*}
% %------------------------------
Figure~\ref{fig:trainloss} shows the average loss of the training and validation sets as a function of training epoch. As can be seen, the loss decreases as a result of the network learning to better represent the data features. 
%------------------------------
%------------------------------
% %------------------------------
% \begin{figure}[ht!]
% \vskip 0.2in
% \begin{center}
%  \centerline{\includegraphics[width=\columnwidth]{FBPoOSEMoDeep_speed.pdf}}
% \caption{The average reconstruction time (single image) and average rRMSE in the test set for the different reconstruction methods, showing that DeepPET outperforms both FBP and OSEM in terms of timing, whilst OSEM has the best rRMSE. Errorbars are standard deviations.}
% \label{fig:losspeed}
% \end{center}
% \vskip -0.2in
% \end{figure}
% %------------------------------
%------------------------------
\begin{figure}[ht!]
\vskip 0.2in
\begin{center}
\centerline{\includegraphics[width=\columnwidth]{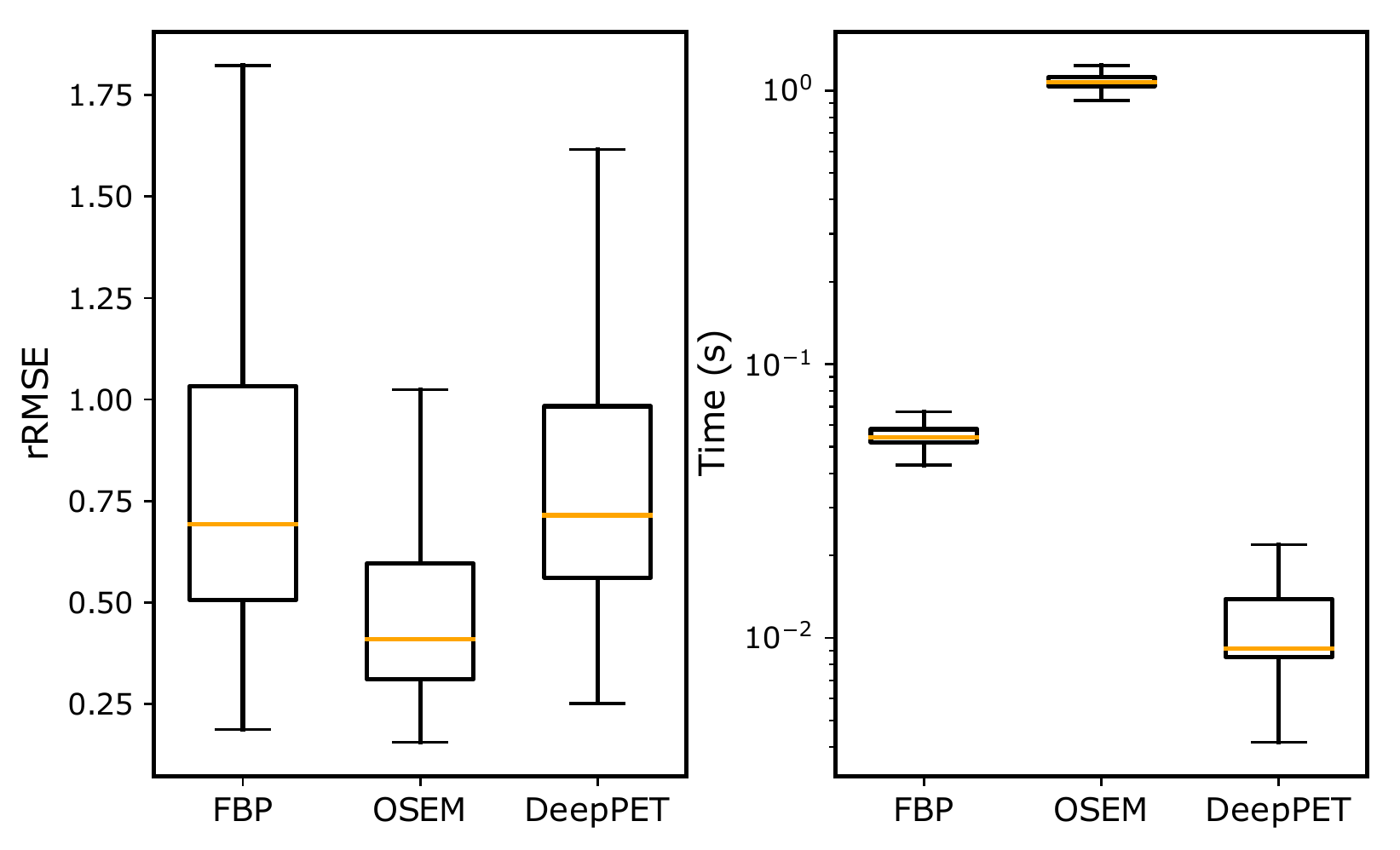}}
\caption{The average rRMSE and the average reconstruction time (single image) in the test set for the different reconstruction methods, showing that DeepPET outperforms both FBP and OSEM in terms of timing, whilst OSEM has the best rRMSE.}
\label{fig:losspeed}
\end{center}
\vskip -0.2in
\end{figure}
%------------------------------
The average reconstruction time per image in the test set, together with the average rRMSE using FBP, OSEM and DeepPET are found in Figure~\ref{fig:losspeed}, and example of successful reconstructions from the test set are seen in Figure~\ref{fig:reconsgood}. Figure~\ref{fig:recons} also includes FBP and OSEM reconstructions, and shows results for different input data noise levels.

%------------------------------
%------------------------------
%------------------------------
% \begin{figure*}[ht]\ContinuedFloat
% \vskip 0.2in
% \begin{center}
% \begin{subfigure}[b]{0.99\textwidth}\ContinuedFloat
%    \centerline{\includegraphics[width=1\linewidth]{FIG3_f.pdf}}
%    \caption{}
% \end{subfigure}
% % \centerline{\includegraphics[width=\textwidth]{FIG3.pdf}}
% \caption{\emph{Continued.} Examples from the test set at different noise levels. For (a), (b), and (c):  Top to bottom: low count (high noise) to high count (low noise) projection data. Left to right: PET projection data (network input), ground truth, FBP reconstruction, OSEM reconstruction, and our Deep reconstruction. Images are labeled with average RMSE relative ground truth average.}
% \label{fig:recons2}
% \end{center}
% \vskip -0.2in
% \end{figure*}
%------------------------------
%------------------------------
\begin{figure*}[ht!]
\vskip 0.2in
\begin{center}
\centering
\centerline{\includegraphics[width=0.9\textwidth]{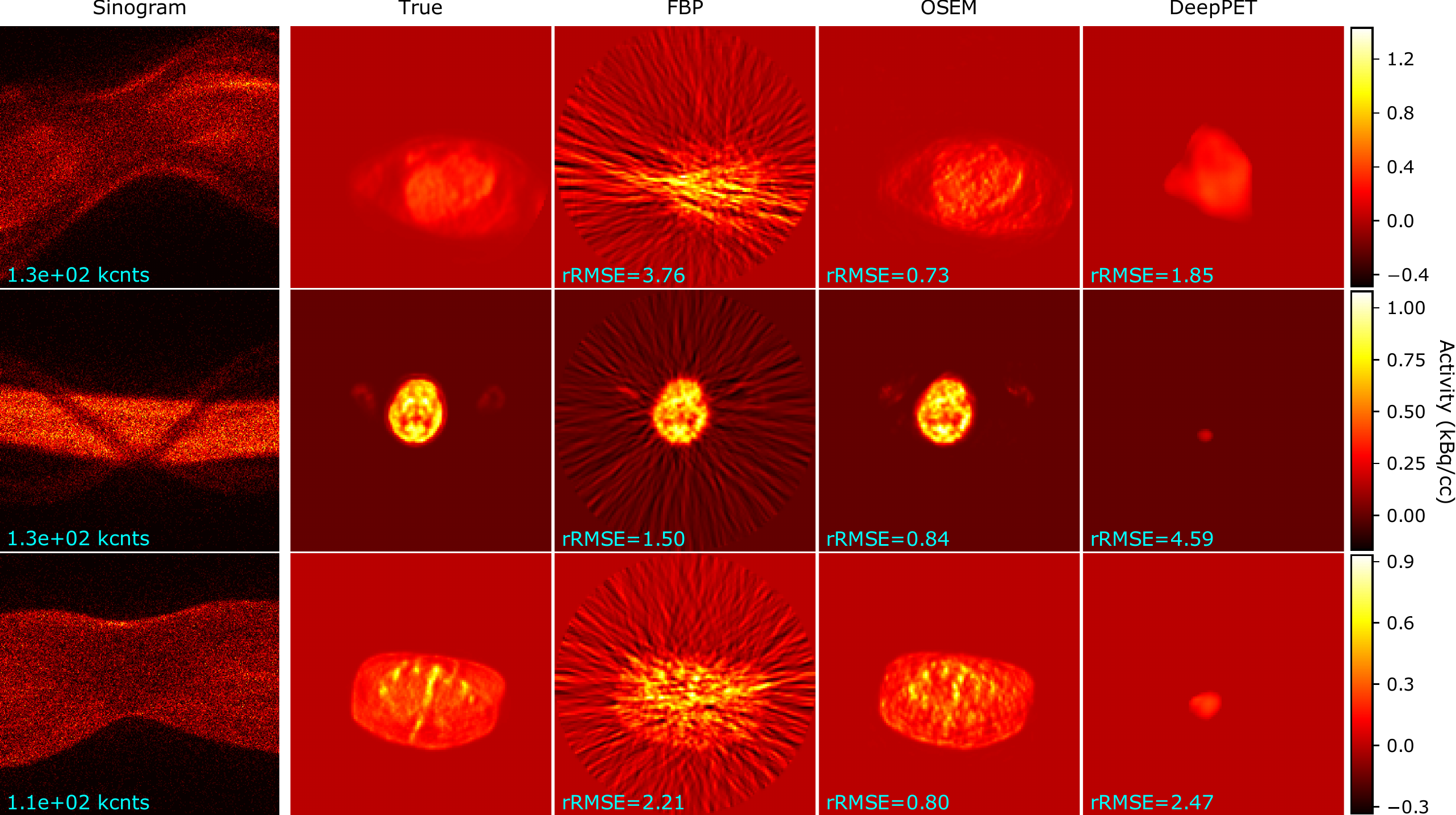}}
\caption{Three examples of very low input data counts, leading to unsuccessful DeepPET reconstructions where most underlying structure is lost. Left to right: PET sinogram (network input), ground truth, FBP, OSEM, and the DeepPET reconstruction. Sinograms are labeled with their total counts, and images with average rRMSE relative ground truth average.}
\label{fig:faulty}
\end{center}
\vskip -0.2in
\end{figure*}
%------------------------------
With an average execution speed of 11~ms per image, DeepPET compares favorable to the conventional methods of FBP at 59~ms and OSEM at 1104~ms, being 104 times faster than OSEM, and 6 times faster than FBP. The average rRMSE of DeepPET is higher at 0.90, compared to FBPs 0.86, and OSEM at 0.50. 
The DeepPET reconstruction favors a slightly smoother solution compared to OSEM, as evident in Figure~\ref{fig:recons}. It should be noted that smoother images, whilst preserving detail, is the main goal of regularizers used in iterative PET reconstruction, as per radiologist preference.
As can be seen, DeepPET reconstruction performs worse for lower count input data (higher noise), as can be expected. This is also true for conventional reconstruction techniques. It was seen however, that for very low counts, DeepPET would at times generate very poor images. Conventional techniques, albeit very poor visual and quantitative quality, typically still manage to reconstruct the gross structure. Examples of poor DeepPET reconstructions can be seen in Figure~\ref{fig:faulty}, where true structure is lost and the result is mostly single or a few blobs. This kind of behavior is however seen in data of lower count than generally found in the clinic (on the order of 100 kcnts). We believe this effect to contribute to the relatively large spread in rRMSE (see Figure~\ref{fig:losspeed}). 

%%%%%%%%%%%%%%%%%%%%%%%%%%%%%%%%%%%%%%%%%%%%%%%%%%%%%
\section{Discussion}
\label{sec:discussion}
One major benefit with PET over many other modalities is that it is inherently quantitative. Hence, the network input and output, though differing in size and structure (as they come from different data domains: sinograms vs. images), are related to one anther, where pixel units go from sinogram counts (registered photon pairs) on the input side, to image pixels in Bq/cc as output.
Furthermore, due to the quantitative nature of the data, we chose not to do any normalization of the input data, since the scaling of the input should decide the output scaling.  
In addition, the spatial correlation between neighboring bins (pixels) in the sinogram is not the same as that in the reconstructed image. The use of a convolutional encoder is therefore not as straight forward and intuitive as when working with ordinary image input. Due to memory, time, and overfitting limitations, a fully connected network on the other hand is very difficult or even infeasible for large 2D data due to the huge number of network weights. As an example, a single fully connected layer taking one sinogram of 288$\times$381 to a 128$\times$128 image requires almost 2 billion weights. 

As shown in the results, DeepPET was 104 times faster than standard OSEM and faster than FBP, where DeepPET only requires one pass to reconstruct an image from sinogram data, whereas traditional techniques require multiple iterations. This would be even more pronounced with regularized reconstruction due to the larger number of iterations typically used. We note that the time savings of DeepPET versus iterative will likely grow larger with the increased use of regularizers. 

%%%%%%%%%%%%%%%%%%%%%%%%%%%%%%%
\subsection{Limitations}
\label{sec:limitations}
One current limitation of the work presented is the use of simulated as opposed to real clinical data. This was the chosen methodology since it allowed us access to ground truth images. Simulated images are based on simplifications and assumptions of the underlying processes, the risk being that the trained network will learn those properties (e.g. approximate randoms as a uniform count over the sinogram) instead of real ones. The more accurate the simulation method is, the lower the risk. Monte Carlo simulations would provide a more accurate representation of the sinogram data, but due to the need of huge datasets ($>$100,000), this could not be achieved within a reasonable time as such simulations typically take days, weeks or even months per image. We instead used PETSTEP, which has been validated against Monte Carlo methods and proved to give realistic results, which gives us confidence that the methodology still holds and is valid for model comparison.

Since we used already reconstructed clinical OSEM images as our ground truth input to PETSTEP, they were inherently noisy with lower resolution already prior to PET acquisition simulation. The CED network will learn based on features included in the training dataset, and hence it is likely that using sharper ground truth images in turn would yield a network better able to produce sharp images.  

We did not include comparisons to any regularized iterative technique in this work, only analytical FBP and unregularized OSEM. Although this is somewhat of a limitation, it also highlights one of the main drawbacks with those approaches. Since there are no general automatic penalty weight selection schemes, it is very difficult to reconstruct on the order of 100,000 images of different anatomies and vastly varying noise levels without user oversight and input. This is the main reason for not including such reconstructions.

We chose to use RMSE as a final metric to compare reconstructed images to ground truth with. Although this is one of the most commonly used image quality metrics, it has limitations. It is generally beneficial for smoother images, even if some detail is lost. Although not included in this study, other metrics could be useful to evaluate resulting image quality.

%%%%%%%%%%%%%%%%%%%%%%%%%%%%%%%
\subsection{Future work}
\label{sec:future}
The deep PET reconstruction method proposed here can be extended to 3D by substituting the 2D convolutions and upsampling steps in the model by their 3D versions. It should be noted that this largely increases the complexity of the network, as well as the time and computer memory needed for training. Furthermore, although this work focuses on PET, the methodology presented is also valid for other types of tomographic data, CT being the most relevant example. CT data is much less noisy than PET, and has higher spatial resolution, making it a very suitable candidate for the approach presented here. 

One next step is to include regularizers within the loss function for training. For example, the same penalties controlling smoothness/sharpness used in conventional regularized iterative reconstruction can be added to the loss function here. Although no effects of overfitting were observed at this point, network weight sparsity can also be enforced.

%%%%%%%%%%%%%%%%%%%%%%%%%%%%%%%%%%%%%%%%%%%%%%%%%%%%%
\section{Conclusions}
\label{sec:concolutions}
To the best of our knowledge this paper presents the first systematic study using a deep learning model that is capable of directly reconstructing PET images from sinogram data.
The major contributions of this work are: (i) We proposed a novel encoder--decoder architecture for PET sinogram data. (ii) On average, DeepPET (time=11~ms) reduces the reconstruction speed over the conventional (OSEM=1.10~s) by a factor of $>$100.
We are confident that our approach shows the potential of deep learning in this domain and is the beginning of a new branch of tomographic PET image reconstruction. Ultimately the gain in quality and speed should lead to higher patient throughput, as well as faster diagnoses and treatment decisions and thus to better care for cancer patients.

%We demonstrated that our proposed network was able to reconstruct images from PET projection data in a few hundredths of a second while {yielding images of higher quality than FBP and higher or comparable to OSEM.

%%%%%%%%%%%%%%%%%%%%%%%%%%%%%%%%%%%%%%%%%%%%%%%%%%%%%
\section*{Acknowledgments}
\noindent
This research was funded in part through the NIH/NCI Cancer Center Support Grant P30 CA008748. The authors are grateful for the generous computational support given by the Warren Alpert foundation. Dr. Thomas J. Fuchs is a founder, equity owner, and Chief Scientific Officer of Paige.AI.


% \bibliography{example_paper}
% \bibliographystyle{icml2018}
\bibliography{References_My_Publications-2018_ICML}
\bibliographystyle{icml2018_ida}
%%%%%%%%%%%%%%%%%%%%%%%%%%%%%%%%%%%%%%%%%%%%%%%%%%%%%

\end{document}